\documentclass{article}
\usepackage{arxiv}
\pdfoutput=1
\usepackage[dvipsnames]{xcolor}
\usepackage[hidelinks]{hyperref}
\usepackage{graphicx}
\usepackage{listings}
\usepackage{amsfonts}
\usepackage{amsmath}
\usepackage{url}
\usepackage{capt-of} 
\usepackage{multirow}
\usepackage{placeins} 
\usepackage{algorithm,algpseudocode}
\usepackage{booktabs}
\usepackage{pgfplots}
\DeclareUnicodeCharacter{2212}{−}
\usepgfplotslibrary{groupplots,dateplot}
\usetikzlibrary{patterns,shapes.arrows}
\pgfplotsset{compat=newest}
\usepackage[textwidth=2cm]{todonotes}

\newcommand{\graphmdlplus}{\textsc{Graph\-\mbox{MDL+}}}
\newcommand{\kgmdl}{\mbox{\textsc{KG-MDL}}}
\newcommand{\KILL}[1]{}
\newtheorem{definition}{Definition}
\newcommand{\vlab}{V_{\cal L}}
\newcommand{\edges}{E_{\cal L}}
\newcommand{\univint}{L_{\mathbb{N}}}
\newcommand{\preq}{L_{preq}}
\newcommand{\blue}[1]{\textcolor{MidnightBlue}{#1}}
\newcommand{\purple}[1]{\textcolor{Plum}{#1}}

\definecolor{CustomGreen}{rgb}{0,.417,.080}
\newcommand{\green}[1]{\textcolor{CustomGreen}{#1}}

\title{\kgmdl{}: Mining Graph Patterns in Knowledge Graphs with the MDL Principle}
\date{September 2023}
\author{%
	Francesco Bariatti\\
	LIACS, Leiden University\\
	Leiden, The Netherlands\\
	\texttt{f.bariatti@liacs.leidenuniv.nl}\\
	\And
	Peggy Cellier\\
	Univ Rennes, INSA Rennes, CNRS, Inria, IRISA -- UMR 6074\\
	F35000 Rennes, France\\
	\texttt{peggy.cellier@irisa.fr}
	\And
	Sébastien Ferré\\
	Univ Rennes, CNRS, Inria, IRISA -- UMR 6074\\
	F35000 Rennes, France\\
	\texttt{sebastien.ferre@irisa.fr}
}

\hypersetup{
pdftitle={\kgmdl{}: Mining Graph Patterns in Knowledge Graphs with the MDL Principle},
pdfsubject={cs.AI},
pdfauthor={Francesco Bariatti, Peggy Cellier, Sébastien Ferré},
pdfkeywords={Knowledge Graphs, Pattern Mining, Graph Mining, Minimum Description Length},
}

\begin{document}

\maketitle

\begin{abstract}
Nowadays, increasingly more data are available as knowledge graphs (KGs).
While this data model supports advanced reasoning and querying, they remain difficult to mine due to their size and complexity.
Graph mining approaches can be used to extract patterns from KGs.
However this presents two main issues.
First, graph mining approaches tend to extract too many patterns for a human analyst to interpret (pattern explosion).
Second, real-life KGs tend to differ from the graphs usually treated in graph mining: they are multigraphs, their vertex degrees tend to follow a power-law, and the way in which they model knowledge can produce spurious patterns.
Recently, a graph mining approach named \graphmdlplus{} has been proposed to tackle the problem of pattern explosion, using the \emph{Minimum Description Length} (MDL) principle.
However, \graphmdlplus{}, like other graph mining approaches, is not suited for KGs without adaptations.
\\
In this paper we propose \kgmdl{}, a graph pattern mining approach based on the MDL principle that, given a KG, generates a \emph{human-sized} and \emph{descriptive} set of graph patterns, and so in a parameter-less and anytime way.
We report on experiments on medium-sized KGs showing that our approach generates sets of patterns that are both small enough to be interpreted by humans and descriptive of the KG.
We show that the extracted patterns highlight relevant characteristics of the data: both of the schema used to create the data, and of the concrete facts it contains.
We also discuss the issues related to mining graph patterns on knowledge graphs, as opposed to other types of graph data.
\end{abstract}


\section{Introduction}

With the growing amount and diversity of data available as knowledge graphs (KGs), there is an increasing need for analyzing them and making sense of their contents.
A first layer of analysis is provided by schemas (RDFS), ontologies (OWL), and constraints (SHACL and ShEx).
These are logical statements expressing absolute truths about the data and usually need to be produced by human experts.
In this paper we are interested in knowledge intermediate between raw facts and logical statements, namely \emph{frequent patterns}, i.e. recurring structures in data.
Such frequent structures can be indicators of regularities in the data, which may in turn be expressed as schemas or constraints, allowing the user to improve the modeling of the KG; or can help optimising query evaluation by highlighting different archetypes of entities~\cite{extended_characteristics_sets}.
Also, since patterns are extracted from raw facts, they can highlight the presence of errors in the data or in the modeling when they do not conform with the expected schema.

\emph{Data mining} is the process of discovering interesting patterns in data~\cite{from_data_mining_to_kdd}.
First applied to transactional data, where transactions and patterns are sets of items, it was later extended to sequential data (e.g., logs, texts)~\cite{agrawal_mining_sequential}, and to graph data (e.g., molecules, text corpora) \cite{survey_frequent_subgraph_mining}.
\emph{Graph pattern mining} was found much more difficult than for transactions and sequences, due to the intrinsic complexity of graph algorithmics.

Existing graph mining algorithms cannot easily be applied to KGs because they were designed for very different kinds of graphs: large collections of small undirected simple graphs (e.g. molecules).
KGs, on the contrary, generally have a single connected component, and they are directed multi-graphs.
Another major difference is that in KGs, node degrees tend to follow a power law~\cite{kg_powerlaw}.
All those differences make graph pattern mining in KGs a very challenging problem that has not been sufficiently addressed so far.
Many existing works applying data mining to KGs have addressed the problem by converting graph data into transactions in order to lower the complexity~\cite{swapriori,swarm,bobed_rdf_evolution}.
A few approaches have been proposed to mine rules from KGs~\cite{amie3,anyburl}.
These are relevant because the rules use a graph pattern as the body (and generally a single triple pattern as the head).
However, the shape of graph patterns is strongly limited, generally to chain patterns.

Pattern mining ---and in particular graph mining--- generally suffers from the large number of extracted patterns.
Several kinds of approaches have been proposed to tackle this problem (e.g., constraints or condensed representations)~\cite{survey_frequent_subgraph_mining}.
Among them, the methods based on the \emph{Minimum Description Length} (MDL) principle showed that it is possible to drastically reduce the number of patterns by selecting a small set of descriptive patterns~\cite{mdl_survey}.
The MDL principle~\cite{mdl_grunwald,mdl_rissanen} comes from the domain of information theory, and states that the \emph{model} that describes the data the best is the one that compresses the data the best, i.e. compression is used as a metric to evaluate models, with sets of patterns used as models in pattern mining.
Several methods based on the MDL principle have been proposed to tackle the number of patterns issue for graph mining~\cite{subdue,mdl_motifs,vog,megs,graphMDLplus}.
However, among them only \graphmdlplus{}~\cite{graphMDLplus} allows to generate patterns without restriction on their shape.
Even so, none of them is directly applicable to KGs.
To our knowledge, only one MDL-based method, called  KGist~\cite{kgist}, has been proposed to target KGs specifically, extracting sets of rules describing the data. However, the shapes of the rules are restricted to rooted trees.

The contribution of this paper is threefold.
First, we propose \kgmdl{}, a MDL-based method to extract a \emph{human-sized} set of \emph{descriptive} patterns from KGs and labeled multi-graphs in general.
It is based on the \graphmdlplus{} approach, but its MDL-related definitions have been revisited to adapt to KGs.
The approach is parameter-free and anytime, and does not put any restriction on the shape of the extracted patterns.
Second, we discuss important distinctions between KGs and the graphs that are usually encountered in graph mining.
Even though both are called ``graphs'', KGs have many peculiarities that must be overcome in order to bring the two domains together.
Finally, we evaluate our approach experimentally on medium-sized KGs, showing that it is effective at extracting a small amount of descriptive patterns.
We also show that the extracted patterns provide useful insights about both the schema used to create the KG, and the raw facts contained in it.
Because of that, they can be a useful asset for exploring unknown KGs, refining ontologies, and introducing constraints.

The rest of this paper is structured as follows:
Section~\ref{sec:related_work} presents related work on graph mining and KG mining.
Section~\ref{sec:preliminaries} presents preliminary notions and the notations that we use in the rest of the paper.
In Section~\ref{sec:kgmdl} we present our contributions, and in Section~\ref{sec:experimental_evaluation} we evaluate it on real-life datasets both from a quantitative and qualitative point of view.
Section~\ref{sec:conclusion} concludes.

\section{Related Work}
\label{sec:related_work}

\paragraph{Graph mining.}
The first graph mining approaches that have been proposed~\cite{survey_frequent_subgraph_mining} were \emph{complete approaches}, i.e. with the aim to extract \emph{all} possible patterns that follow a certain acceptance rule, usually a minimum frequency of appearance (e.g. gSpan~\cite{gspan}, Gaston~\cite{gaston}).
A common drawback of this kind of approaches is that they extract a large amount of patterns, which makes human analysis difficult.
Approaches have been proposed to reduce the amount of extracted patterns using  constraints or condensed representations, with mitigated results~\cite{survey_frequent_subgraph_mining}.

More recently, graph mining approaches have been proposed that use the \emph{Minimum Description Length} (MDL) principle~\cite{mdl_grunwald,mdl_rissanen} to more drastically reduce the amount of extracted patterns, and to obtain graph patterns that are descriptive of the data.
In this family of methods, Subdue~\cite{subdue} can extract patterns with any shape, but its encoding entails a loss of information about the graph structure.
It is possible to know which patterns appear in the data, but not \emph{how they connect to each other}, which is a major disadvantage for analysis.
VoG and MeGS~\cite{vog,megs} are methods targeted at summarizing unlabeled graphs that have no such loss of information, but only consider a pre-defined vocabulary of pattern shapes. While this vocabulary can be expanded, doing so requires to have prior knowledge about the type of structures that can be observed in the data.
The approach proposed in~\cite{mdl_motifs} allows to extract \emph{motifs} from unlabeled graphs, which are similar in the idea to patterns. This approach interprets MDL encodings as probability distributions, which are needed to select motifs. However it only applies to unlabeled graphs, and only when an \emph{a priori hypothesis} about their shape exists.
Finally, \graphmdlplus{}~\cite{graphMDLplus} can mine patterns with arbitrary shapes from both directed and undirected graphs. The approach introduces the notion of ``ports'' to avoid loss of information about the graph structure.
However, all these approaches can not be applied directly to KG mining.
This is in part because the type of graphs that they support does not encompass labeled multi-graphs (which knowledge graphs are), and in part because ---as we detail in Section~\ref{sec:kgmdl:kg_representation}--- KGs have peculiarities that differentiates them from the graphs found in a classic graph mining context, namely their highly skewed degree distribution and the presence of connections between semantically unrelated elements.

\paragraph{Knowledge graph mining.}
Some data mining approaches have been proposed to target knowledge graphs specifically.
Some approaches such as SWApriori~\cite{swapriori}, SWARM~\cite{swarm} and the one proposed in~\cite{bobed_rdf_evolution} convert the KG to a set of transactions to lower the complexity of the task, and then apply classic itemset-mining algorithms to it.
However, this conversion limits the shape of the patterns that can be extracted.
Most other KG mining approaches, such as AMIE~\cite{amie3}, AnyBURL~\cite{anyburl}, RuDik~\cite{rudik}, and KGist~\cite{kgist} (which also uses the MDL principle) mine association rules whose bodies are graph patterns.
However, in these approaches the shape of the patterns is also limited, often to rooted trees, or requires a minimum connectivity on the vertices in the pattern.
Another common drawback of KG mining methods is that they are often developed for (and evaluated on) a specific task (e.g. link prediction, error correction), and do not perform other tasks, in contrast with the open-minded goal of data exploration that is central to data mining~\cite{kg_refinement_survey}.

\section{Preliminaries}
\label{sec:preliminaries}

In this section we define some notions that are used extensively in the rest of the paper.

\subsection{Graphs}
\label{sec:preliminaries:labeled_graphs}

\begin{figure}[t]
	\centering
	\begin{minipage}[b]{0.65\textwidth}
		\centering
		\includegraphics[width=\textwidth]{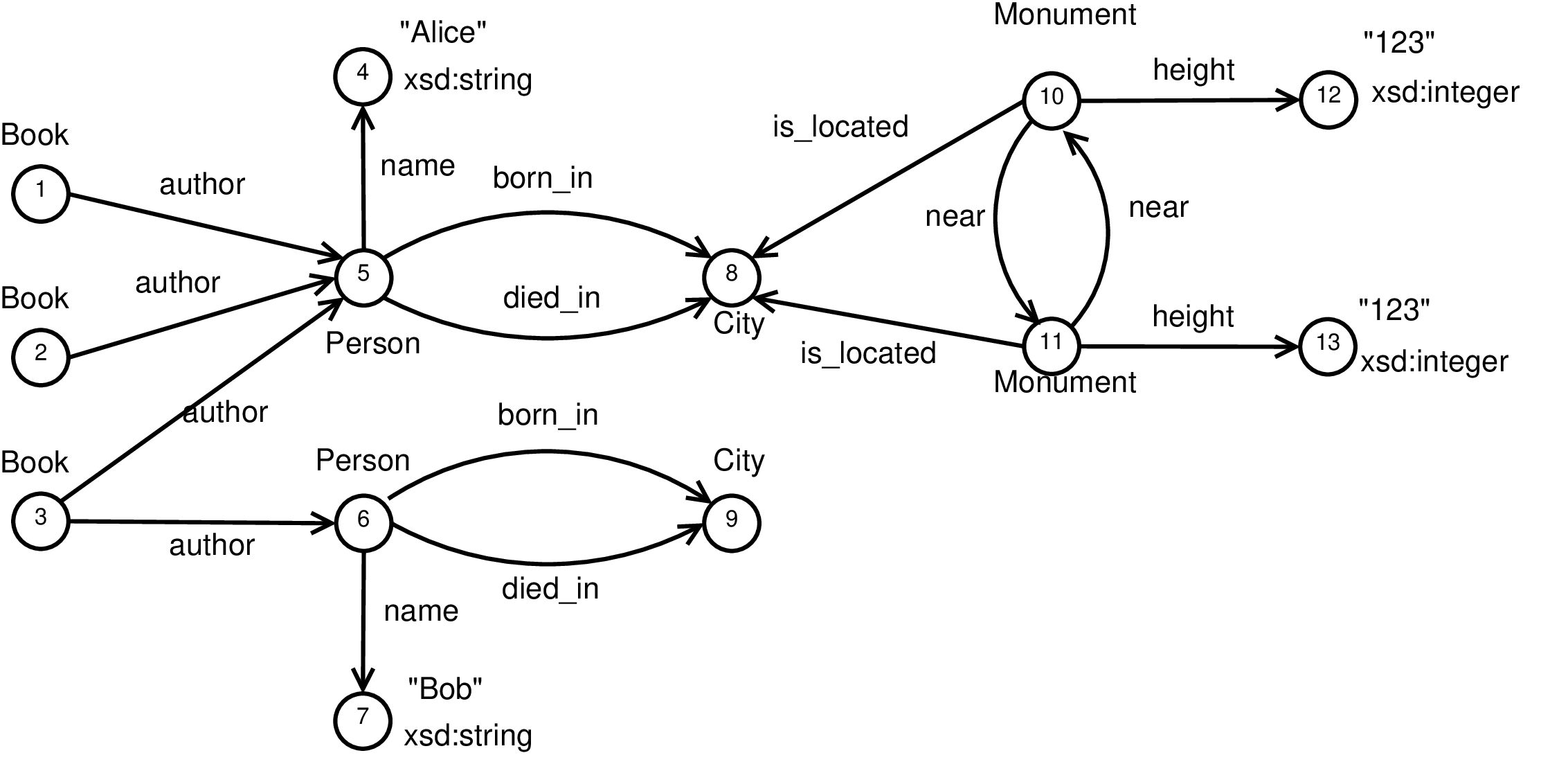}
		\caption{An example of a directed multi-graph, used as example data throughout the paper.}
		\label{fig:example_multigraph}
	\end{minipage}
	\hfil
	\begin{minipage}[b]{0.3\textwidth}
		\centering
		\includegraphics[width=0.7\textwidth]{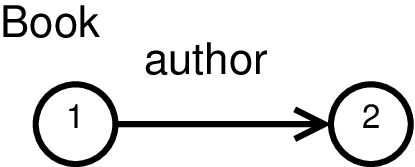}
		\vspace{3\baselineskip}
		\caption{An example pattern which has four occurrences in the graph of Fig~\ref{fig:example_multigraph}.}
		\label{fig:example_pattern}
	\end{minipage}
\end{figure}

In this paper, we define \emph{graphs} using the following structure:

\begin{definition}
A \emph{graph} $G = (V, \vlab, \edges)$ over a set of \emph{label symbols} ${\cal L}$ is a data structure composed of a set of \emph{vertices}~$V$, a set of \emph{vertex labels} \mbox{$\vlab \subseteq V \times \mathcal{L}$}, and a set of \emph{edge labels} ---or \emph{labeled edges}--- \mbox{$\edges \subseteq V \times V \times \mathcal{L}$}.
\end{definition}

\noindent
This definition encompasses all graphs that are usually found in the literature: it allows for both directed and undirected graphs (depending on whether $(u,v,l) \in \edges$ is equivalent to $(v, u ,l) \in \edges$ or not); it allows vertices to have any number of labels; and finally it allows for multi-graphs, i.e. graphs where multiple edges with different labels can exist between two vertices.
Note that in Section~\ref{sec:kgmdl:kg_representation} we discuss how to represent KGs (defined as set of triples) with this definition, and important considerations to take into account in doing so when targeting a graph mining context.
Fig.~\ref{fig:example_multigraph} shows an example graph, with 13 vertices (named 1 to 13), 17 vertex labels, and 16 edge labels, from a set of 16 label symbols. This graph will be used as example data throughout this paper.
Note that many existing graph mining approaches are limited in the kind of graphs that they can accept. For example, most accept only simple graphs, i.e. graphs that contain no self loops (edges from a vertex to itself) and a maximum of one edge between any pair of vertices. Many also require vertices to have exactly one label.

\begin{definition}[Pattern occurrence]
Let  $G^P$ (the ``pattern'') and $G^D$ (the ``data'') be graphs. An \emph{occurrence} ---or \emph{embedding}--- of $G^P$ in $G^D$ is an injective function \mbox{$\varepsilon \in V^P \to V^D$} such that:
(1) $(v, l) \in \vlab^P \implies (\varepsilon(v), l) \in \vlab^D$,
(2) $(v_1, v_2, l) \in \edges^P \implies (\varepsilon(v_1), \varepsilon(v_2), l) \in \edges^D$.
\end{definition}

\noindent
Informally, a pattern occurrence is a function that maps all vertices of the ``pattern'' to some of the vertices of the ``data''. It conveys the information that the structure of the pattern can be found at some specific point in the data.
Note that this definition represents the occurrences under \emph{isomorphism}, meaning that each pattern vertex must be mapped to a \emph{different} data vertex. This is the definition commonly used in graph mining, whereas the SPARQL query language uses \emph{homomorphism} (i.e. two different variables in a SPARQL graph pattern can be bound to the same RDF node)\footnote{It is possible to use isomorphism in SPARQL graph patterns by adding a FILTER clause expressing that all variables must be pairwise distinct.}.
Fig.~\ref{fig:example_pattern} shows an example pattern that has four occurrences in the graph of Fig.~\ref{fig:example_multigraph}: three occurrences mapping vertex 1 of the pattern to respectively vertices 1, 2, or 3 of the data and vertex 2 of the pattern to vertex 5 of the data; and one occurrence mapping vertex 1 of the pattern to vertex 3 of the data and vertex 2 to vertex 6.

\subsection{The MDL Principle}
\label{sec:preliminaries:mdl}

The \emph{Minimum Description Length} (MDL) principle~\cite{mdl_grunwald,mdl_rissanen} comes from the domain of information theory.
It allows to identify the model ---from a family of models--- that best describes some data.
It has been successfully used in pattern mining to select a small and descriptive set of patterns on some data, by considering sets of patterns as possible models for the data~\cite{mdl_survey}.

\begin{definition}[MDL principle]
Given a family of models ${\cal M}$ and some data $D$, the best model $M \in {\cal M}$ to describe the data is the one minimizing the \emph{description length} $L(D, M) = L(M) + L(D|M)$, where $L(M)$ is the description length of the model, and $L(D|M)$ is the description length of the data encoded with the model.
\end{definition}

\noindent
The MDL principle is generic for any kind of data and models.
Each MDL-based approach must devise its own definitions for the different description lengths. However, a few primitives are commonly used~\cite{mdl_introduction}.
We list here the ones used in this paper (note that in the whole paper, $\log$ indicates the base 2 logarithm):

\begin{itemize}
	\item An element~$x$ chosen with uniform probability from a set~${\cal X}$ has a description length of $\log(|{\cal X}|)$ bits.

	\item $k$ elements chosen with uniform probability from a set ${\cal X}$ have a description length of $\log(\binom{|{\cal X}|}{k})$ bits.

	\item An element~$x \in {\cal X}$ appearing $\#x$ times in data has a description length of $L_{\cal X}(x) = -\log(\frac{\#x}{\sum_{x_i \in {\cal X}} \#x_i})$ bits. This is often called a \emph{prefix code}~\cite{mdl_introduction,mdl_grunwald}.

	\item An integer $n \in \mathbb{N}$ with no known upper bound has a description length of $\univint(n)$ bits, where $\univint$ is a \emph{universal integer encoding}~\cite{elias_encodings}\footnote{In our implementation we use the \emph{Elias delta encoding}, shifted by 1 so that it can encode 0.}.

	\item A sequence $S$ of elements drawn from a set ${\cal X}$ has a description length of $\preq({\cal X}, S)$, where $\preq$ is a \emph{prequential plugin code}~\cite{mdl_grunwald} (see Appendix~\ref{sec:appendix:prequential} for details)\footnote{In our implementation we initialize the usage of elements in the prequential encoding to $\epsilon = 0.5$, according to literature.}.
\end{itemize}

The act of associating a description length to an element is called \emph{encoding}.
Note that in MDL-based approaches what matter are the \emph{theoretical} description lengths as stated above, and the actual encodings do not need to be materialised~\cite{mdl_grunwald}.
In particular, it is common and acceptable for description lengths to be non-integer amounts.
However, the MDL principle still requires encodings to be (at least theoretically) reversible: meaning that no information should be lost. This is the case for all primitives listed above.

\section{\kgmdl}
\label{sec:kgmdl}

In this section, we formally define our approach, \kgmdl{}, which is an adaptation of \graphmdlplus{}~\cite{graphMDLplus} to KGs.
First in Section~\ref{sec:kgmdl:kg_representation} we discuss the need for a pre-processing stage before a KG can be used in a pattern mining task, showing that the standard graph representation of KGs presents some issues in a graph mining context, and we propose an alternative (and reversible) representation.
In Section~\ref{sec:kgmdl:mdl_encoding} we present the MDL encoding and related description lengths that are used in \kgmdl{} in order to evaluate sets of patterns.
Then, in Section~\ref{sec:kgmdl:algorithm} we present the search algorithm used to search for the best set of patterns according to the MDL criterion.
Finally, in Section~\ref{sec:kgmdl:in_practice} we discuss some issues that arise when applying our approach to real-life KGs, due to the differences between those KGs and the graphs usually found in the graph mining domain, and we present our solution to avoid them.

\subsection{Knowledge Graphs Representation for Pattern Mining}
\label{sec:kgmdl:kg_representation}

\begin{figure}
	\centering
	\includegraphics[width=0.9\textwidth]{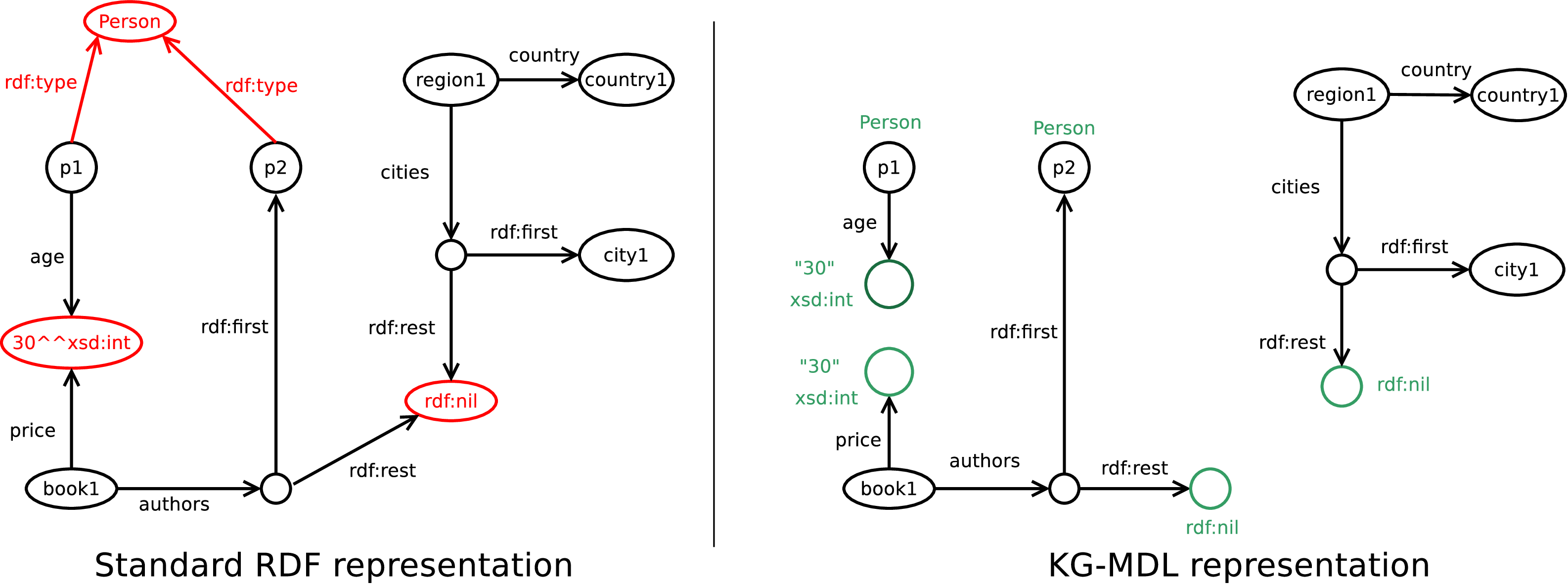}
	\caption{Left: a KG represented using the usual RDF graph representation. Right: the same KG represented following \kgmdl{}'s mapping. The differences between the two representations are emphasized with colours.}
	\label{fig:kg_as_labeled_graph}
\end{figure}

The RDF standard defines a way to convert a KG expressed as a collection of triples to a graph representation.
It consists in creating a vertex for each element appearing as subject or object, and adding an edge for each triple ---labeled with the triple's predicate--- connecting the vertex corresponding to the subject to the vertex corresponding to the object.
An example of such a representation is shown on the left of Fig.~\ref{fig:kg_as_labeled_graph}.

While we could use this representation directly (as the graph definition given in Section~\ref{sec:preliminaries:labeled_graphs} covers it), we show here that this standard representation is not suited when the resulting graph is to be used in a graph mining context.
A first problem that arises with this representation is about literals: in Fig.~\ref{fig:kg_as_labeled_graph} we can see that in the standard representation a single vertex is used for all instances of the integer literal 30. As a consequence, vertices \verb|book1| and \verb|p1| are neighbours in the graph, even though the value ``30'' represents a price for the former and an age for the latter.
This is not a problem when answering queries (hence why the standard uses this representation), but becomes a problem if the graph is used to generate patterns: we would see spurious patterns such as ``a book that costs as many dollars as a person is old in years'', which relate entities just because they happen to have the same numerical value for semantically different quantities.
An equivalent problem happens with the \verb|rdf:nil| entity, which is used to mark the end of \emph{all} lists. If not treated carefully, patterns can be generated relating semantically unrelated lists, just because of them being lists.
Finally, \verb|rdf:type| edges can also generate spurious patterns by making all entities of the same type neighbours in the graph. In graph mining, attributes such as the type of an entity are instead represented as vertex labels.

In conclusion, before mining patterns, \kgmdl{} converts the data KG to what is shown on the right of Fig~\ref{fig:kg_as_labeled_graph}: the object of triples with predicate \verb|rdf:type| is added as a label to the vertex corresponding to the triple's subject; for each triple having a literal object, a new \emph{distinct} vertex is created for the literal, labeled with the literal's type and value (and language for language-tagged strings); for each triple having \verb|rdf:nil| as subject, a new distinct vertex is created, labeled \verb|rdf:nil|. For other triples, an edge is added between the vertices corresponding to the subject and object.
Note that this conversion is reversible: it is always possible to convert the graph back to the original KG\footnote{On the condition of storing which labels were originally types and which were literal datatypes and values, which is an easy task.}.
In particular, the patterns extracted from the graph can be converted to SPARQL queries that ---when executed on the original KG--- will give mappings corresponding to the patterns' occurrences.
For example, the pattern of Fig.~\ref{fig:example_pattern} can be converted to \verb|SELECT ?x1 ?x2 WHERE { ?x1 a <Book> ; <author> ?x2 .}|.

\subsection{MDL Encoding for Knowledge Graphs}
\label{sec:kgmdl:mdl_encoding}

In this section we present the MDL encoding used by our approach for computing the MDL description length (see Section~\ref{sec:preliminaries:mdl}) needed to compare possible sets of patterns.
This encoding is based on the one used by \graphmdlplus{}~\cite{graphMDLplus}, from which it adopts the same high-level structure.
However, since \graphmdlplus{} considers both the data and the patterns to be simple graphs ---i.e. graphs that have no edges from a vertex to itself and no more than one edge between each pair of vertices--- its encoding does not support the full expressiveness of knowledge graphs. Instead, new description lengths must be defined.

The encoding used by our approach is composed of two parts: the \emph{code table} (corresponding to MDL's ``model'') and the \emph{rewritten graph} (corresponding to MDL's ``data encoded with the model'').

\subsubsection{The Model: Code Table}

\begin{figure}
	\centering
	\includegraphics[width=0.45\textwidth]{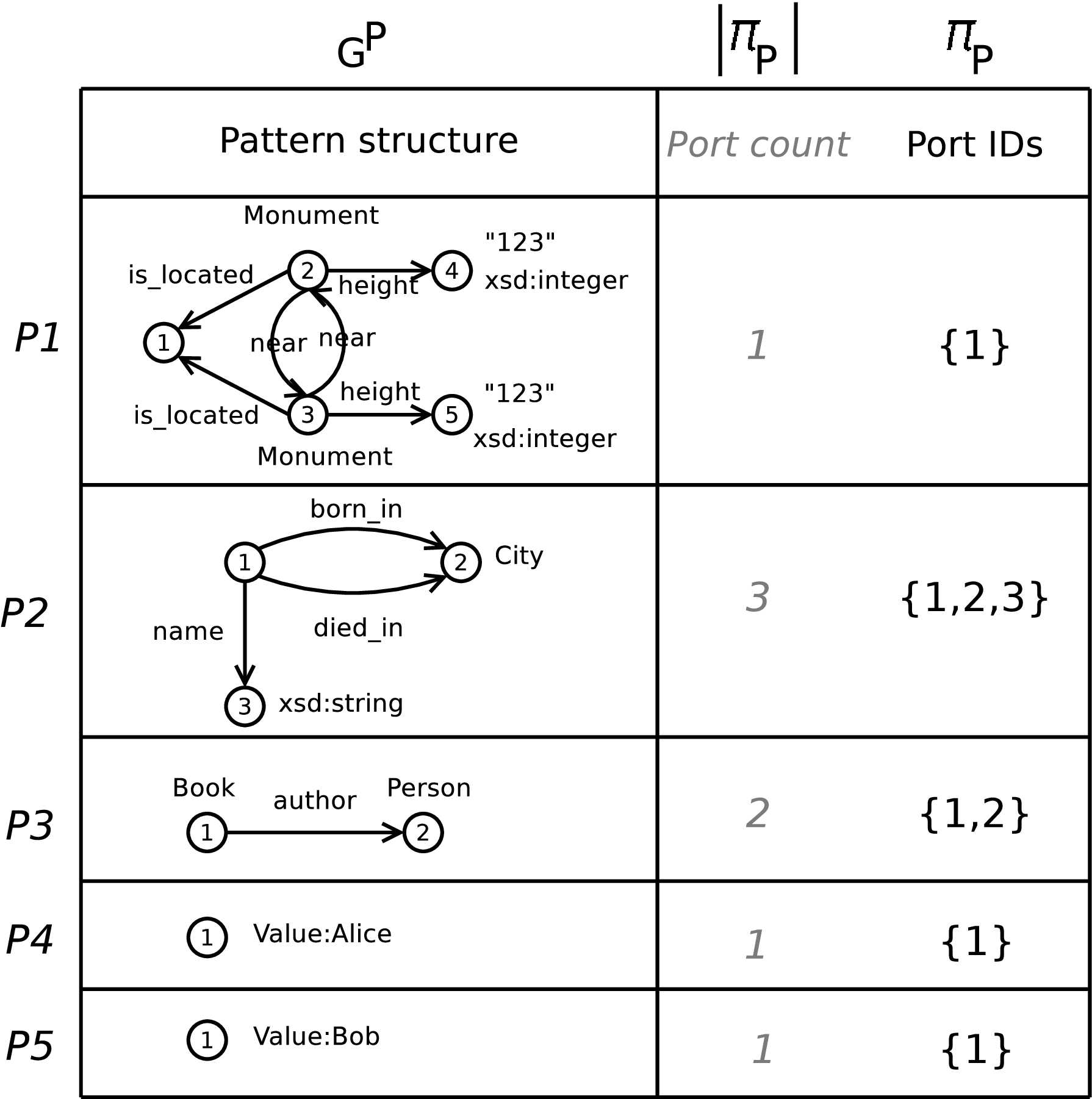}
	\caption{A \kgmdl{}'s \emph{code table}.}
	\label{fig:ct}
\end{figure}

\kgmdl{}'s model is called a \emph{code table} (CT): it is a collection of patterns, with some associated information that will be used when encoding the data with those patterns.
Fig.~\ref{fig:ct} shows an example code table that can be used to encode the graph of Fig.~\ref{fig:example_multigraph}.
Each row of the code table is a tuple $P = (\blue{G^P}, \green{\Pi_P})$ representing the characteristics of a pattern: its \blue{graph structure $G^P = (V^P, \vlab^P, \edges^P)$} and its set of \green{\emph{ports}~$\Pi_P \subseteq V^P$}.
The notion of ports, first introduced in~\cite{graphMDL}, is key for encoding graph data with graph patterns. The ports of a pattern are the vertices that the pattern's occurrences share with other pattern occurrences when the CT is used to encode the data (see below).
For example, the first row of Fig.~\ref{fig:ct} represents the pattern ``two monuments of height 123 that are near to each other and located in the same place''.
This pattern has only one port, its vertex~1, meaning that when the CT is used to encode the data, occurrences of this pattern can only be connected to other pattern occurrences by overlapping on their vertex~1.

\subsubsection{The Encoded Data: Rewritten Graph}

\begin{figure}
	\centering
	\includegraphics[width=\textwidth]{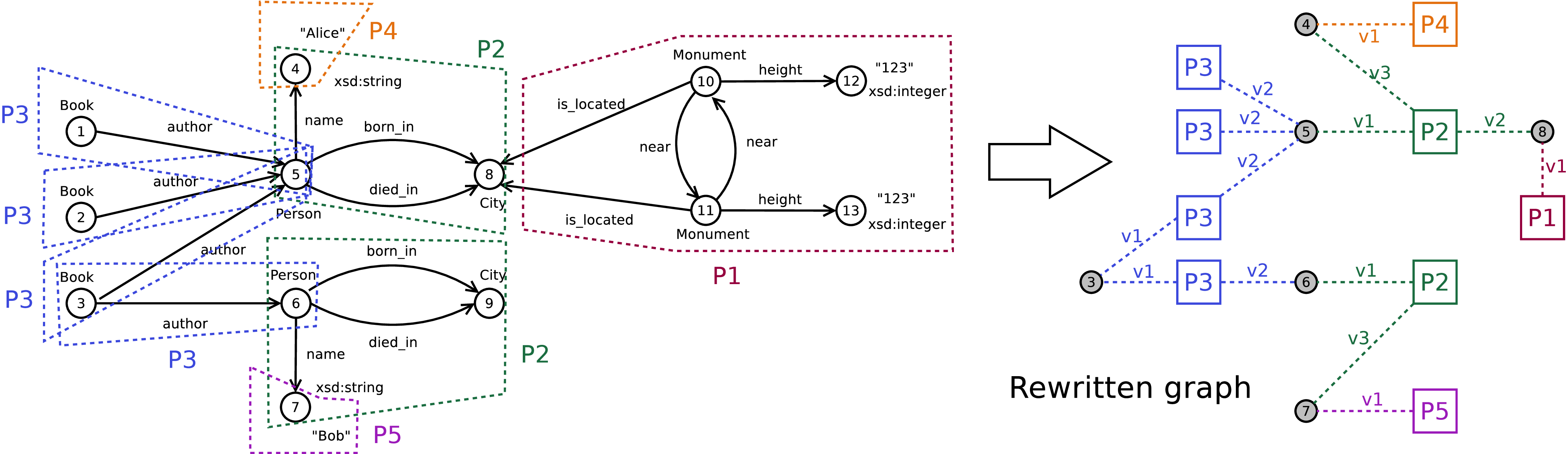}
	\caption{How the patterns of the code table of Fig.~\ref{fig:ct} can be used to encode the data graph of Fig.~\ref{fig:example_multigraph}, and the resulting rewritten graph.}
	\label{fig:rewritten_graph}
\end{figure}

In \kgmdl{}, the ``data encoded with the model'' is represented by what is called a \emph{rewritten graph}.
The rewritten graph ---first introduced in~\cite{graphMDL}--- is a data structure allowing to represent a data graph as a composition of pattern occurrences, without any loss of information about the structure of the data.
Fig.~\ref{fig:rewritten_graph} shows a rewritten graph corresponding to the encoding of the data graph of Fig.~\ref{fig:example_multigraph} with the patterns in the code table of Fig.~\ref{fig:ct}.

\begin{definition}[from~\cite{graphMDL}, with adapted notation]
A rewritten graph~$G^R = (\blue{(V^R_{emb}, V^R_{port})}, \green{\vlab^R}, \purple{\edges^R})$ is a special graph where: $\blue{V^R_{emb}}$ is the set of \emph{embedding vertices}, $\blue{V^R_{port}}$ is the set of \emph{port vertices}, $\green{\vlab^R = V^R_{emb} \to CT}$ a function associating CT patterns to embedding vertices, and $\purple{\edges^R \subset V^R_{emb} \times V^R_{port} \times \biguplus_{P \in CT} \Pi_P}$ a set of edges between embedding vertices and port vertices, labeled with pattern ports.
Such edges are subject to the condition $(v_e, v_p, \pi) \in \edges^R \implies \pi \in \Pi_P \text{ with } P = \vlab^R(v_e)$, i.e. edges can only be labeled with port identifiers for the pattern associated to their embedding vertex.
\end{definition}

\noindent
The rewritten graph is composed of two disjoint sets of vertices: each \emph{\blue{embedding vertex}}~$v_e \in V^R_{emb}$ (squares in Fig.~\ref{fig:rewritten_graph}) represents an occurrence of a pattern of the CT, which describes part of the data graph.
Each embedding vertex has one \green{label} telling to which pattern~$P \in CT$ it corresponds.
\emph{\blue{Port vertices}}~$v_p \in V^R_{port}$ (circles in Fig.~\ref{fig:rewritten_graph}) represent those data vertices which are ``covered'' by at least two pattern occurrences associated to embedding vertices.
Finally, \emph{\purple{edges}}~$(v_e, v_p, \pi) \in \edges^R$ can connect embedding vertices to port vertices, telling that a specific embedding vertex uses a specific port vertex. These edges have a label~$\pi \in \Pi_P$ ---where $\Pi_P$ are the possible ports of the pattern associated to $v_e$--- telling which of the pattern's vertices is mapped to the data vertex corresponding to $v_p$.
For example, in the top-left of Fig.~\ref{fig:rewritten_graph} we see that part of the data (around vertices 1 and 5) is described as an occurrence of pattern~$P_3$, which shares its second vertex (vertex 5 of the data) with other pattern occurrences.
In the rewritten graph, this is represented with an embedding vertex labeled~$P_3$ connected with an edge to a port vertex that represents vertex~5 of the data, and the edge is labeled~$v_2$ to indicate that it corresponds to vertex~2 of the pattern. Other pattern occurrences that also use the same data vertex are represented as embedding vertices also connected to the same port vertex.
Data vertices that are not shared between pattern occurrences do not appear in the rewritten graph.

The rewritten graph has two uses: firstly it is needed for applying the MDL principle, as the ``data encoded with the model'' is part of the MDL description length. But it can also be given to the user as a companion to the selected set of patterns: in this way the user can not only analyze the selected patterns, but also how they connect and interact with each other.
For example, in Fig.~\ref{fig:rewritten_graph} there are four occurrences of pattern $P_3$ (``a book has an author who is a person''), and for three of them their vertex 2 is mapped to the same port vertex, surfacing the knowledge that most of the books share the same author in the data graph.

\subsubsection{Description Lengths}
\label{sec:kgmdl:description_lengths}

In this section we define the MDL description length that \kgmdl{} uses to compare different sets of patterns.
The MDL description length is composed of two parts (see Section~\ref{sec:preliminaries:mdl}): the model length~$L(M)$ and the encoded data length~$L(D|M)$. In \kgmdl{} the model is a code table, and the encoded data is a rewritten graph.
All formulas are shown in Fig.~\ref{fig:kgmdl:dl}.

In this section we use the following shorthand notations:
1) \mbox{$\{\edges^R|P\} = \{(v_e, v_p, \pi) \in \edges^R \mid \vlab^R(v_e) = P\}$} is the set of rewritten graph edges that start from an embedding vertex of pattern $P$,
2) for a set of tuples~$X = \{(a_1, b_1), \dots\}$ we write $(-, b) \in X$ to mean the \emph{sequence} composed of all values $b_i$ appearing in $X$, in an arbitrary but determined order. This notation can be generalised to tuples of any size and to any of their components.

\begin{figure}[t]
	\begin{align*}
		& \text{\underline{\small Code table description length}}\\
		&L(M) = L(CT) = \sum_{P \in CT}
			\blue{
				\overbrace{L(G^P)}^{\text{Pattern structure}}
			} +
			\green{
				\overbrace{
					\underbrace{\log(|V^P|+1)}_{\text{port count} |\Pi_P|} +
					\underbrace{\log(\binom{|V^P|}{|\Pi_P|})}_{\text{port identities}}
				}^{\text{ports } \Pi_P}
			}
		\\
		& \blue{L(G^P)} =
		\underbrace{\univint(|V^P|)}_{\text{vertex count}} +
		\underbrace{\log(|{\cal L}|)}_{\text{label symbols count}} +
		\sum_{\{l \in {\cal L} \mid l~\text{is in}~P\}} \left [
			\underbrace{L_{\cal L}(l)}_{\text{symbol identity}} +
			\underbrace{\univint(\#occ)}_{\text{occurrences count}} +
			\underbrace{\log(\binom{|V^P|^n}{\#occ})}_{\text{occurrences}}
		\right ]\\
		& \text{\underline{\small Rewritten graph description length}}\\
		&L(D|M) = L(G^R) =
			\blue{\overbrace{\univint(|V^R_{emb}|)}^{V^R_{emb}}} +
			\blue{\overbrace{\log(|V^D|+1)}^{V^R_{port}}} +
			\green{\overbrace{L_{preq}(CT, (-, P) \in \vlab^R)}^{\vlab^R}} \quad +
			\\
			& \qquad \purple{ \overbrace{
				\sum_{P \in CT} \left [
					\underbrace{\sum_{v_e \in \{\edges^R|P\}}\log(|\Pi_P|+1)}_\text{edge sources} +
					\underbrace{L_{preq}(\Pi_P, (-,-, \pi) \in \{\edges^R |P\})}_\text{edge labels}
				\right] +
				\underbrace{L_{preq}(V^R_{port}, (-,v_p,-) \in \edges^R)}_\text{edge destinations}
			}^{\edges^R}}
	\end{align*}
	\caption{MDL description length used by \kgmdl{} to evaluate sets of patterns.}
	\label{fig:kgmdl:dl}
\end{figure}

\paragraph{Code table description length.}
The description length of a code table corresponds to the sum of the description length of each of its rows.
For each row the information that needs to be encoded is: the \blue{pattern's structure $G^P$}, and the \green{pattern's ports $\Pi_P$}.

In order to encode the \blue{pattern's structure}, we first need to encode the number of vertices~$|V^P|$ of the pattern. Since a pattern can have any number of vertices, we use a universal integer encoding $\univint$ for this information.
Then, we describe all the labels that appear in the pattern.
For that, we first encode how many different label symbols can be found in the pattern: since a pattern can contain between one\footnote{A pattern with zero labels would not make sense, as it would correspond to a structure with no edges and no vertex labels.} and as many different symbols as there are in the data (i.e. ${\cal L}$), this information can be encoded with $\log(|{\cal L}|)$~bits.
Then, for each label symbol present in the pattern, we encode: its identity, how many occurrences it has in the pattern, and which ones they are.
In order to indicate which label symbol $l \in {\cal L}$ we are describing, we use a prefix code $L_{\cal L}(l)$ based on the number of occurrences that each label has in the data graph.
Then, we encode how many occurrences ($\#occ$) this symbol has in the pattern with a universal integer encoding.
Finally, we indicate what those occurrences are: in a pattern with $|V^P|$~vertices, a vertex label has $|V^P|$ possible occurrences, and an edge label has $|V^P|^2$. Thus, we can encode them using $log(\binom{|V^P|^n}{\#occ})$~bits, with $n$ either 1 or 2 depending on the nature of the label symbol.
Note that this encoding does not put any limit on the shape of a pattern, and can encode any graph structure.

Once the pattern's structure has been encoded, encoding the \green{pattern's ports} is simple, as the ports are a subset of the pattern's vertices.
First, the number of ports $|\Pi_P|$ (between 0 and all vertices) is encoded with $\log(|V^P|+1)$ bits, then the choice of which of the vertices are ports is encoded with $\log(\binom{|V^P|}{|\Pi_P|})$ bits.

\paragraph{Rewritten graph description length.}
The description length of a rewritten graph corresponds to the sum of the description length of each of its elements: \blue{the vertices}, \green{the vertex labels}, and \purple{the edges}.

In order to encode \blue{the vertices} of the rewritten graph it suffices to encode how many embedding vertices and how many port vertices there are.
The number of embedding vertices can be anything and is thus encoded with a universal integer encoding, whereas the number of ports is bounded by the number of vertices in the data (at maximum patterns can superpose on all of them) and can therefore be encoded with $\log(|V^D|+1)$~bits.
For the example of Fig.~\ref{fig:rewritten_graph}, this corresponds to $\blue{\univint(9) + \log(14)}$.

In order to encode the \green{vertex labels}, since we encoded the set of embedding vertices and each one of them has one label, we can use a prequential code with elements drawn from the set of patterns in the CT.
For the example of Fig.~\ref{fig:rewritten_graph}, this corresponds to $\green{L_{preq}(\{P_1, P_2, P_3, P_4, P_5\}, \langle P_3, P_3, P_3, P_3, P_4, P_2, P_2, P_5, P_1 \rangle)}$

We encode \purple{the edges}~$(v_e, v_p, \pi) \in \edges^R$ one element at a time. First, for each pattern we encode the number of edges starting from each of its embedding vertices~$v_e$. Since an embedding vertex can use at maximum all ports~$\Pi_P$ of its pattern (and at minimum 0), its number of edges can be encoded with $\log(|\Pi_P|+1)$~bits.
Then, for each pattern we encode the label~$\pi$ of all edges of its embedding vertices (whose number and sources have been encoded in the previous step), which can be done using a prequential code with elements drawn from the set of ports~$\Pi_P$ of the pattern.
For example, for pattern $P_3$ of Fig.~\ref{fig:rewritten_graph}, this corresponds to $\purple{4\log(3) + L_{preq}(\{v_1,v_2\}, \langle v_2,v_2,v_2,v_1,v_1,v_2 \rangle)}$.
Finally, the destinations~$v_p$ of all edges in the rewritten graph can be encoded using a prequential code with elements drawn from the set of port vertices~$V^R_{port}$.
For the example of Fig.~\ref{fig:rewritten_graph}, this corresponds to $\purple{L_{preq}(\{3,4,5,6,7,8\}, \langle 5,5,5,3,3,6,4,4,5,8,6,7,7,8 \rangle)}$.

In total, the the description length of the rewritten graph of Fig.~\ref{fig:rewritten_graph} is given by:
\begin{align*}
&\blue{\univint(9) + \log(14)}
+ \green{L_{preq}(\{P_1, P_2, P_3, P_4, P_5\}, \langle P_3, P_3, P_3, P_3, P_4, P_2, P_2, P_5, P_1 \rangle)}\\
&\purple{
	+ \underbrace{4\log(3) + L_{preq}(\{v_1,v_2\}, \langle v_2,v_2,v_2,v_1,v_1,v_2 \rangle)}_{P_3}
	+ \underbrace{\log(2) + L_{preq}(\{v_1\}, \langle v_1 \rangle)}_{P_4}
	+ \underbrace{\dots}_{P_2, P_5, P_1}
}\\
&\purple{+ L_{preq}(\{3,4,5,6,7,8\}, \langle 5,5,5,3,3,6,4,4,5,8,6,7,7,8 \rangle)}
\end{align*}

\subsection{Search Algorithm}
\label{sec:kgmdl:algorithm}

\begin{algorithm}[t]
	\begin{algorithmic}[1]
		\Require A data graph $D$
		\State $CT_0 \gets$ initial singleton-only code table from labels in $D$
		\State \Return \Call{Iteration}{$D, CT_0$}
		\State 
		\Function{Iteration}{$D, CT$}
			\State $G^R \gets$ rewritten graph from $D$ and $CT$
			\State ${\cal C} \gets$ candidates generated from pairs of embedding vertices in $G^R$
			\State ${\cal C}_{ranked} \gets {\cal C}$ sorted by heuristic \Comment See~\cite{graphMDLplus}
			\ForAll{$C \in {\cal C}_{ranked}$}
				\State $CT' \gets CT \cup C$
				\If{$L(D, CT') < L(D, CT)$}
					\State \Return \Call{Iteration}{$D, CT'$}
				\EndIf
			\EndFor
			\State \Return $CT$
		\EndFunction
	\end{algorithmic}
\caption{Search algorithm used by \kgmdl{}.}
\label{alg:search_algorithm}
\end{algorithm}

The algorithm that \kgmdl{} uses in order to search for the best possible code table is shown in Alg.~\ref{alg:search_algorithm}. It is the same search algorithm used by \graphmdlplus{}~\cite{graphMDLplus}.
The algorithm starts (line 1) with an initial code table $CT_0$ composed of singleton patterns only, i.e. patterns having either a single vertex with a single label, or a single edge between unlabeled vertices.
Then it proceeds iteratively: it computes the rewritten graph for the current code table (line 5), and looks at which patterns appear sharing ports in the rewritten graph. Each pair of embedding vertices that are neighbours in the rewritten graph generates a \emph{candidate pattern} corresponding to the merge of the patterns of the two embedding vertices, along their shared ports (line 6).
Such candidates are ordered following a heuristic (line 7), and then one by one the algorithm tries to add them to the current code table (lines 8 and 9): if one is found where the description length of the new code table is smaller than the previous one (which in MDL terms means it is better), a new iteration is started with this new code table (lines 10 and 11).
If no candidate gives a better code table the algorithm stops, returning the best code table found (line 14).

This algorithm is parameterless and anytime: it requires only the data graph as input, and when it is stopped it can return the best code table found so far.
Patterns are generated bottom-up as merges of smaller patterns.
The only difference between the algorithm used by \kgmdl{} and the one used by \graphmdlplus{} resides in the formulas used to compute the MDL description length $L(D,CT)$: \kgmdl{} uses the formulas introduced in Section~\ref{sec:kgmdl:description_lengths}, which can handle multigraphs (and thus KGs).

\subsection{When Theory Meets Practice}
\label{sec:kgmdl:in_practice}

When applying \kgmdl{} on real-life KGs we observed a high memory usage and a high runtime that were not observed when applying \graphmdlplus{} to classical graph mining datasets.
Based on our analysis, those problems arose because the basic implementation of the search algorithm would compute for each pattern in the CT all of its occurrences in the data, and go through them all in order to compute the rewritten graph.
In real-life KGs some pattern emerged that have an exponentially high number of possible occurrences.
An example of such patterns is ``N elements located in the same country'': if the data graph contains hundreds of countries with hundreds of elements each, the number of possible occurrences of this pattern is exponentially high.
This type of structure seems very typical of KGs, which often have predicates used to ``list'' elements, which results in some vertices with a particularly high number of edges around them, following a power-law~\cite{kg_powerlaw}.
In contrast, in the datasets usually used in graph mining there is often an upper bound on the degree of a vertex: e.g. when mining molecules, the maximum number of edges is the maximum number of bonds for an atom, which is a small value.
This is yet another difference between KGs and classic graph mining datasets, which must be catered for when applying data mining techniques to KGs.

Our solution to limit this problem has two parts.
First, we avoid computing (and storing) \emph{all} occurrences of each pattern, as only few of them are required to create the rewritten graph.
Indeed, while the rewritten graph allows pattern to overlap on vertices, it forbids edge overlaps~\cite{graphMDL}, therefore a large number of possible occurrences can be discarded.
Second, we add the possibility to specify a maximum time to spend on each code table row when creating the rewritten graph.
We call this parameter the \emph{row cover timeout}.
This allows to enforce that a fairer share of time is allocated to all patterns, avoiding that some complex patterns slow down the whole computation.
This is important in the context of an anytime approach such as \kgmdl{}, since time is an important resource and as such it is crucial to avoid spending too much time on a small part of the search space.
We evaluate this solution experimentally in Section~\ref{sec:experimental_evaluation}.

\section{Experimental Evaluation}
\label{sec:experimental_evaluation}

In this section we evaluate \kgmdl{} on five real-life datasets.
We first evaluate the approach quantitatively, and show that \kgmdl{} is able to extract a small set of patterns which covers the data efficiently, and minimize the MDL description length.
Then, we present some of the extracted patterns, showing that they convey relevant information about both the schema of the data, and the concrete facts contained in the data.

\subsection{Settings and Datasets}

\begin{table}
	\caption{Characteristics of the datasets used in the experiments.}
	\label{tab:dataset_characteristics}
	\centering
    \begin{tabular}{@{}lrrrrr@{}}
		\toprule
		& & \multicolumn{4}{c}{\kgmdl{} graph modeling} \\
		\cmidrule{3-6}
        \multicolumn{1}{c}{Dataset} & \#triples & $|V|$ & $|\vlab|$ & $|\edges|$ & $|{\cal L}|$ \\
        \midrule
        SemanticBible-NTNames	& 4k	& 2k	& 5k  	& 3k	& 1k \\
        Lemon-dbpedia-en 		& 30k	& 14k	& 19k	& 17k	& 2k \\
        Mondial-europe			& 50k	& 22k	& 28k	& 34k	& 3k \\
        Taaable					& 138k	& 82k	& 140k	& 115k	& 17k \\
        Mondial					& 168k	& 74k	& 95k	& 114k	& 8k \\
        \bottomrule
    \end{tabular}
\end{table}

In order to evaluate \kgmdl{} we developed a prototype in Java, which is available as a git repository\footnote{\url{https://gitlab.inria.fr/fbariatt/graphmdl}}.
Experiments were run on a single core (our prototype is not multi-threaded) of an Intel Xeon E5-2630 v3 CPU, and had a RAM limit of 32GB.

We conducted the experiments on five datasets.
The {\bf Lemon-dbpedia-en} dataset\footnote{\url{https://lemon-model.net/lexica/dbpedia_en/}} is a lexicalisation of the most common DBpedia classes and properties using the ``Lemon'' ontology. It contains information such as \emph{``The term \emph{island} is a common noun whose sense is the \texttt{dbo:Island} class''}. In this dataset ---because of the way it is used--- we considered the \verb|partOfSpeech| predicate equivalent to the \verb|rdf:type| predicate.
The {\bf Mondial} dataset\footnote{\url{https://www.dbis.informatik.uni-goettingen.de/Mondial/}} contains geographical information, such as \emph{``Mt Everest is a mountain in the Himalayas''}. The {\bf Mondial-europe} dataset is the subset covering Europe only.
Since these two datasets contain a large number of numerical values, we removed the concrete values and encoded them with a vertex labeled by the type of the value only (e.g. \texttt{xsd:Integer}).
The {\bf SemanticBible-NTNames} dataset\footnote{\url{http://www.semanticbible.com/ntn/ntn-overview.html}} is a semantic web representation of named entities in the Bible with associated information, such as {\em ``Adam is a man, spouse of Eve, parent of Abel, Cain and Seth''}.
The {\bf Taaable} dataset\footnote{We obtained the data from the authors of~\cite{taaable}.} corresponds to an RDF representation of the Wikitaaable semantic wiki~\cite{taaable}, containing recipe and food information such as \emph{``Bread is a baked good made using 2 teaspoons of salt''}.

The characteristics of the datasets are presented in Table~\ref{tab:dataset_characteristics}. For each dataset we report the number of triples in its original RDF representation and the number of vertices, vertex labels, edges, and the total number of label symbols in their \kgmdl{} representation (presented in Section~\ref{sec:kgmdl:kg_representation}).
For example the Lemon dataset is composed of 30 thousand triples, which in \kgmdl{} correspond to 14 thousand vertices having a total of 19 thousand vertex labels, connected by 17 thousand edges; in total there are 2 thousand unique label symbols in the dataset.

The results of our experiments are available online on the GraphMDL Visualizer website\footnote{\url{http://graphmdl-viz.irisa.fr/}}, which interactively displays for each dataset global statistics as well as all the extracted patterns.

\subsection{Quantitative Evaluation}

\begin{table}
	\caption{Results of running \kgmdl{} on the different datasets for 8h with a row cover timeout of 200ms, 500ms and 1s.
    Only the timeout yielding the lowest description length is reported for each dataset.}
	\label{tab:quantitative_evaluation}
	\centering
	\resizebox{\textwidth}{!}{
	\begin{tabular}{@{}lcrcccr@{}}
		\toprule
				&				&				& \multicolumn{2}{c}{Data labels described}	&												&					\\
				& 				&				& \multicolumn{2}{c}{by patterns with}		& Compression ratio								& Time to reach \\
		\cmidrule{4-5}
        Dataset & Row cover timeout 		& \#patterns	& $\ge 2$ labels		& $\ge 10$ labels	& $L\% = \frac{L(CT_{kgmdl}, D)}{L(CT_0, D)}$ 	& 95\% of L\% gain \\
        \midrule
        NTNames 		& 200ms	& 234	& 86.6\%	& 44.8\%	& 50.8\%	& 6 min \\
        Lemon 			& 200ms	& 168	& 94.9\%	& 87.5\%	& 20.3\%	& 4 min \\
        Mondial-europe	& 500ms	& 118	& 92.0\%	& 79.1\%	& 34.1\%	& 1 h 45 min \\
        Taaable			& 200ms	& 93	& 83.8\%	& 76.1\%	& 41.0\%	& 2 h 25 min \\
        Mondial			& 500ms	& 78	& 90.2\%	& 55.4\%	& 34.6\%	& 2 h 32 min \\
        \bottomrule
    \end{tabular}
	}
\end{table}

Table~\ref{tab:quantitative_evaluation} presents the results obtained when running \kgmdl{} on the different datasets.
For each dataset three experiments were run with a row cover timeout of respectively 200ms, 500ms, and 1s. Each experiment was given 8 hours (the Lemon experiments actually terminated after 4h having finished their exploration).
We only report in the table for each dataset the experiment that gave the best (i.e. the lowest) description length at the end of the 8 hours~(we evaluate the impact of the row cover timeout further along this section).
The first column of the table reports the dataset, and the second one reports the timeout for that experiment.

\paragraph{Number of generated patterns.} The third column of Table~\ref{tab:quantitative_evaluation} (\#patterns) reports the number of patterns in the best code table found by \kgmdl{}.
We observe that on all datasets, even on the biggest ones, the number of generated patterns is in the lower hundreds.
We argue that it is plausible for a human analyst to comprehend such an amount of patterns, and observe them in order to get an idea of the structure of the data.
Note that classic graph mining algorithms would tend to extract tens of thousands or hundreds of thousands patterns for datasets of this size.
Thus, we can conclude that \kgmdl{} is effective at its goal of generating a \emph{small and human-sized} set of patterns.

\paragraph{Labels described by patterns.}
The fourth column of Table~\ref{tab:quantitative_evaluation} reports how many of each dataset's labels are described in the rewritten graph by instances of the patterns generated by \kgmdl{}, the other labels being described by the trivial one-label patterns of the $CT_0$.
We observe that on all datasets more than 80\% of the labels are described by a pattern, which shows that \kgmdl{} patterns are descriptive enough to cover most of the data.
When looking at the data in detail, we notice that most of the non-covered labels correspond to concrete literal values (e.g. ``123'' or ``Alice'').
This is consistent with the idea of finding patterns to describe the \emph{structure} of the data, without caring about the concrete values of the different properties.

For the fifth column of Table~\ref{tab:quantitative_evaluation} we take a stricter limit of only considering the data labels covered by patterns with at least ten labels in their structure. We observe that despite limiting to the occurrences of these large patterns, the amount of data labels described remains high.
This shows that the data contains large regularities that \kgmdl{} is able to capture, which would probably interest the analyst.

\paragraph{Description length.} The sixth column of Table~\ref{tab:quantitative_evaluation} reports the compression ratio $L\% = \frac{L(CT_{kgmdl}, D)}{L(CT_0, D)}$, corresponding to the ratio between the description length given by the code table found at the end of the experiment and the one given by the initial code table~$CT_0$.
This is the MDL measure that \kgmdl{} tries to minimize.
We observe that the approach is efficient in doing so, obtaining a compression ratio that ranges from 20\% to 50\% depending on the dataset.
From an information theory point of view, that means that \kgmdl{} manages to find a set of patterns that makes it possible to describe the data with a fifth to half as much information than without those patterns.

\begin{figure}
	\centering
	\begin{minipage}[t]{0.4\textwidth}
		\vspace{0pt} 
		\centering
		\resizebox{\textwidth}{!}{
		\input{img/dl_evolution}
		}
		\caption{Evolution of the compression ratio $L\%$ given by the CT found by \kgmdl{} as a function of time. The dotted line of each dataset represents the compression ratio that is 95\% of the way w.r.t. the last one.}
		\label{fig:dl_evolution}
	\end{minipage}
	\hfil
	\begin{minipage}[t]{0.45\textwidth}
		\captionof{table}{Compression ratio $L\%$ on the Mondial dataset for different row cover timeouts (8h experiment).}
		\label{tab:timeout_impact}
		\centering
		\begin{tabular}{@{}cc@{}}
			\toprule
			Timeout per CT row & \multirow{2}{*}{Compression ratio $L\%$} \\
			during search & \\
			\midrule
			100ms & 36.0\% \\
			200ms & 34.6\% \\
			\textbf{500ms} & \textbf{34.6\%} \\
			1s & 37.3\% \\
			5s & 41.3\% \\
			10s & 40.7\% \\
			\bottomrule
		\end{tabular}
	\end{minipage}
\end{figure}

\paragraph{Description length evolution over time.}
We expect that a user would probably not want to wait for 8 hours in order to get a result from their experiments. Since \kgmdl{} is an anytime algorithm, it can be stopped at any point during the search, yielding the best result found so far.
In order to evaluate the impact of stopping the algorithm sooner, we show in Fig.~\ref{fig:dl_evolution} the evolution of description length over time.
Each point of the plot corresponds to one iteration of the search algorithm, i.e. to a code table that is found which is better than the previous one.
We observe that most of the gain is obtained in the first few minutes or hours. For instance, on the Mondial dataset, the compression ratio goes from 100\% (the starting model) to around 40\% in the first hour, and the last 7 hours are spent going from 40\% to 35\%.
On top of that, as time advances the points tend to be more spaced out, as it takes more time to find a code table that manages to improve the previous one.

We report in the last column of Table~\ref{tab:quantitative_evaluation} the time needed in each experiment to attain a compression ratio that is 95\% of the way w.r.t. the full 8 hours experiment.
We observe that for all datasets, \kgmdl{} obtains good results very quickly, which means that the user can usually stop the algorithm early and still obtain good results.

\paragraph{Impact of the row cover timeout.}
In order to evaluate the impact of the row cover timeout parameter described in Section~\ref{sec:kgmdl:in_practice} more in depth, we ran some additional experiments on the Mondial dataset, whose results are shown in Table~\ref{tab:timeout_impact}.
We observe that a timeout too low or too high can negatively impact the final description length (which is also the case for the other datasets of Table~\ref{tab:quantitative_evaluation}, as the best description length is not obtained with the same timeout value for all datasets).
We believe that too low a timeout can stop the computation before some important occurrences are found for some patterns, which makes the algorithm under-estimate useful patterns.
On the other hand, too high a timeout entails the risk of spending so much time on the computation of the occurrences of few complex patterns, that it will limit the amount of different patterns that can be processed in a given time.
In conclusion, the cover timeout is a useful parameter, which can make a difference in the performances of the algorithm.

\subsection{Qualitative Evaluation}
\label{sec:experimental_evaluation:qualitative}

In this section we take a look at the patterns that are selected by \kgmdl{} and we show that they allow to surface knowledge about both the schema of the KG from which they are extracted, and from the actual raw data, conveying knowledge that is not present in the schema.
Note that the patterns are shown here with the graph representation presented in Section~\ref{sec:kgmdl:kg_representation}. However, they can be converted seamlessly to SPARQL queries that can be executed on the original data. In Appendix~\ref{sec:appendix:sparql_patterns} we show the SPARQL queries corresponding to the patterns in this section.

\subsubsection{Schema Knowledge from the Patterns}

\begin{figure}[t]
	\centering
	\begin{minipage}[b]{0.35\textwidth}
		\centering
		\includegraphics[width=\textwidth]{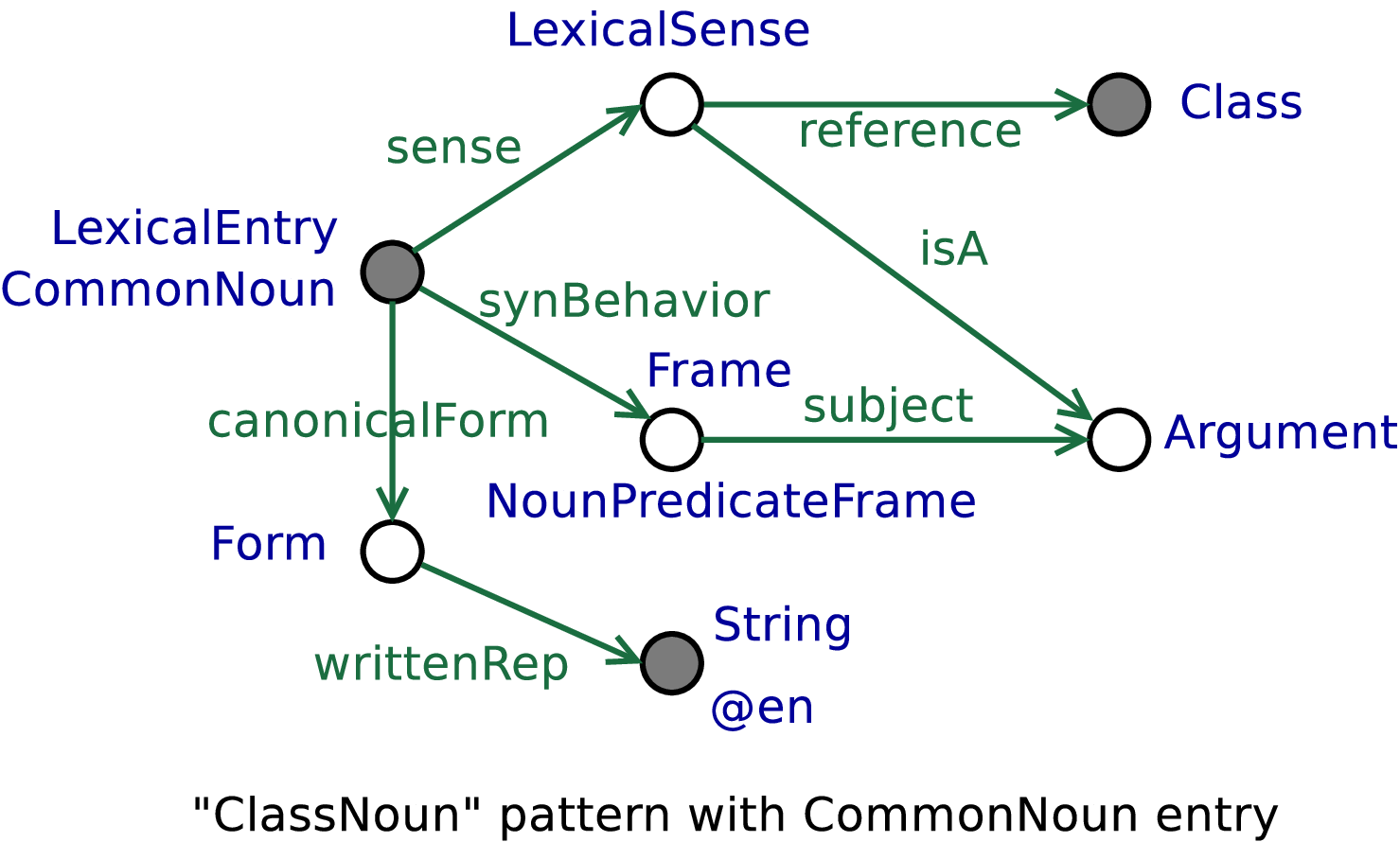}
		\caption{The pattern that appears the most in the \kgmdl{} rewritten graph on the Lemon dataset. Vertices filled in grey are the ports of the pattern.}
	\label{fig:lemon_classnoun}
	\end{minipage}
	\hfil
	\vrule
	\hfil
	\begin{minipage}[b]{0.6\textwidth}
		\centering
		\includegraphics[width=\textwidth]{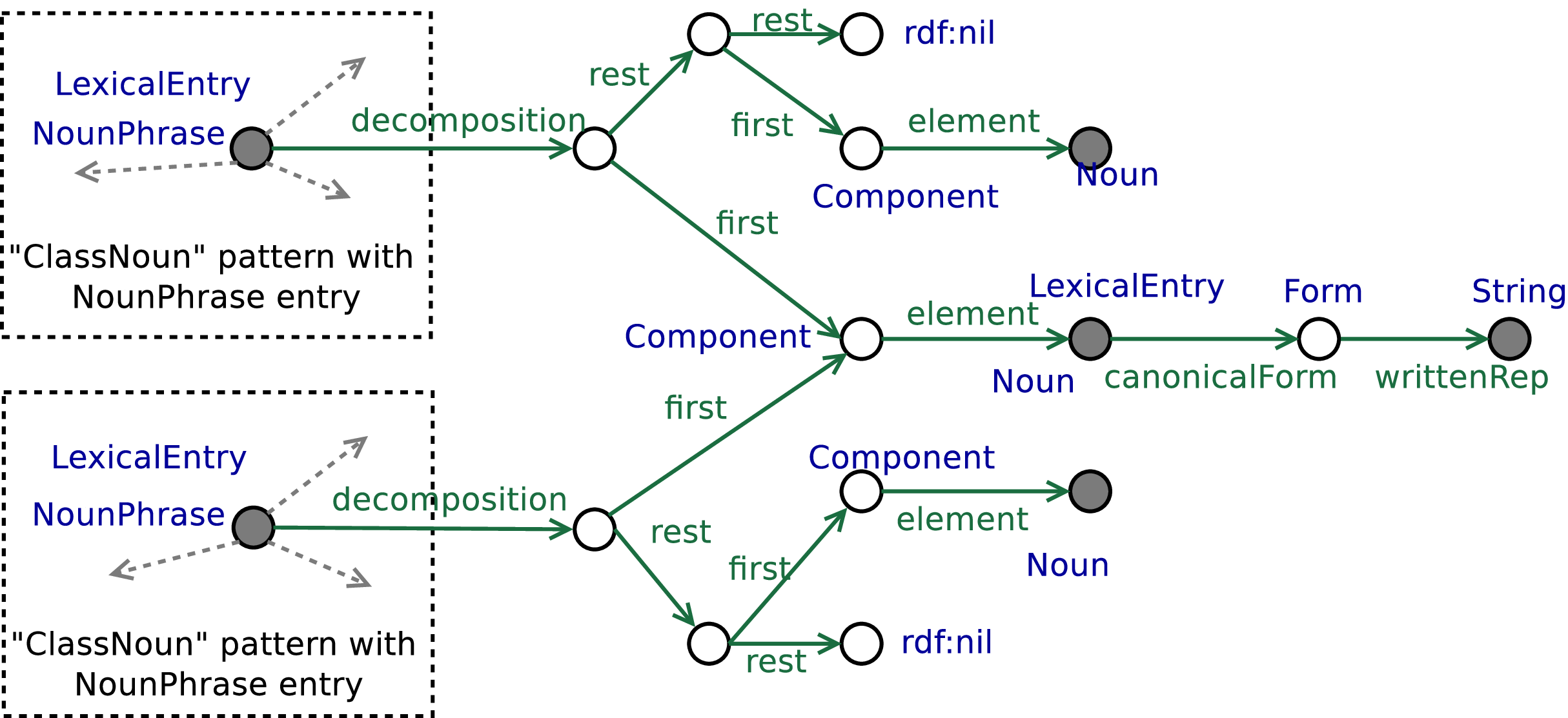}
		\caption{A pattern extracted by \kgmdl{} on the Lemon dataset. The squares on the left correspond to the structure of Fig.~\ref{fig:lemon_classnoun} except for label \mbox{``CommonNoun''} being replaced by label \mbox{``NounPhrase''}.}
		\label{fig:lemon_nounphrases}
	\end{minipage}
\end{figure}

By looking at the github page of the Lemon dataset\footnote{\url{https://github.com/ag-sc/lemon.dbpedia}}, we observe that the dataset is created using some predefined structures called \emph{lemon design patterns}.
\kgmdl{} does not have any knowledge of those design patterns, but all of them (around ten different types of design patterns are used in our dataset) are found in the set of extracted patterns, demonstrating the ability of our approach to identify them as important structures in the data.
Fig.~\ref{fig:lemon_classnoun} shows the pattern that appears the most in the \kgmdl{} rewritten graph on the lemon dataset: it corresponds to the lemon design pattern ``ClassNoun''. This structure describes a lexical entry that is a common noun (e.g. \emph{``lake''}), and whose meaning is a class (e.g. \texttt{dbo:Lake}).
The structure used in Lemon to encode this entry is not trivial: the pattern contains 7 vertices, 10 vertex labels, and 7 edge labels.
The pattern combines three linguistic aspects: lexical (the written representation of the term), syntactic (a noun predicate frame with one argument, the subject), and semantic (the sense is given as a reference to an RDFS class). Syntax and semantics are aligned through the Argument vertex, which links the subject of the frame and the instance of the class.
Only three out of the seven pattern vertices ---filled in grey in the figure--- appear as ports in the rewritten graph: the lexical entry, the class, and the written representation. They correspond to the vertices that differ from one pattern occurrence to another, i.e. they have different additional labels, either vertex labels or edges connecting them to other pattern occurrences. Hence, each pattern occurrence is fully determined by the choice of three data vertices.
This corresponds to how the data is created for the lemon dataset, where only those pieces of information need to be provided per instance, the rest of the structure being created automatically.
In this case, \kgmdl{} allowed to highlight the \emph{schema} of the data: the extracted pattern shows the user a characteristic shape from the data that stems from how the data has been designed.
This shape could, for instance, be converted to a SHACL constraint in order to ensure that a KG corresponds to the correct design.

\paragraph{}
Fig.~\ref{fig:lemon_nounphrases} shows another pattern extracted from the Lemon dataset. Contrary to the previous one, this pattern does not correspond directly to a lemon design pattern (what we could call the ``schema'' of the graph).
It represents a pair of lexical entries of type \mbox{``NounPhrase''} (i.e. terms composed by multiple elements) that share the same first element, but whose meaning are distinct classes, e.g. \texttt{dbo:BloodVessel} and \texttt{dbo:BloodType}, \texttt{dbo:RadioProgram} and \texttt{dbo:RadioStation}, or \texttt{dbo:MountainRange} and \texttt{dbo:MountainPass}.
This patterns tells us that it is common to have terms with different meanings but whose lexical form starts with the same word.
Indeed the pattern has 34 occurrences in the rewritten graph (i.e. it describes 68 lexical entries).
Since this pattern shows how to encode this kind of lexical entries, it corresponds to knowledge that could be present in the schema of the data (the design patterns).
Therefore, in this case \kgmdl{} surfaced knowledge that could become an \emph{addition to the schema}, but was not explicitly present in the schema during the design.

\paragraph{}
Fig.~\ref{fig:ntnames_siblings} shows one of the patterns extracted by \kgmdl{} on the NTNames dataset.
In the OWL description of this dataset {\tt childOf} and {\tt parentOf} are marked as inverse properties, and {\tt siblingOf} is marked as symmetric.
\kgmdl{} generates a pattern that strongly suggests those axioms without having access to the schema itself, just from the data.
The algorithm also suggests an axiom that is not present in the schema of the dataset: when two people have the same parent, they are siblings one of the other.
In this case \kgmdl{} shows in the same pattern both knowledge that is present in the schema of the data, and knowledge that could be an addition to the schema.

\subsubsection{Factual Knowledge from the Patterns}

\begin{figure}[t]
	\centering
	\begin{minipage}[b]{0.4\textwidth}
		\centering
		\includegraphics[width=\textwidth]{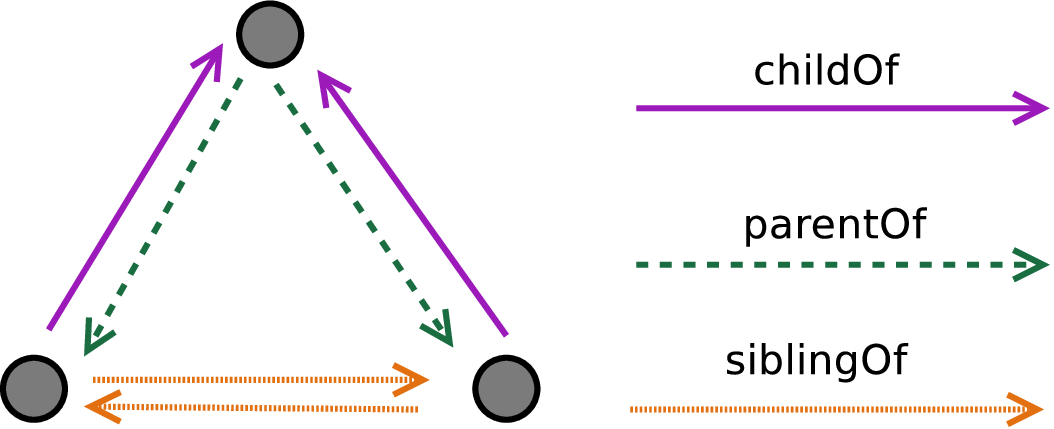}
		\caption{A pattern extracted by \kgmdl{} on the NTNames dataset. Vertices filled in grey are the ports of the pattern.}
		\label{fig:ntnames_siblings}
	\end{minipage}
	\hfil
	\begin{minipage}[b]{0.45\textwidth}
		\centering
		\includegraphics[width=\textwidth]{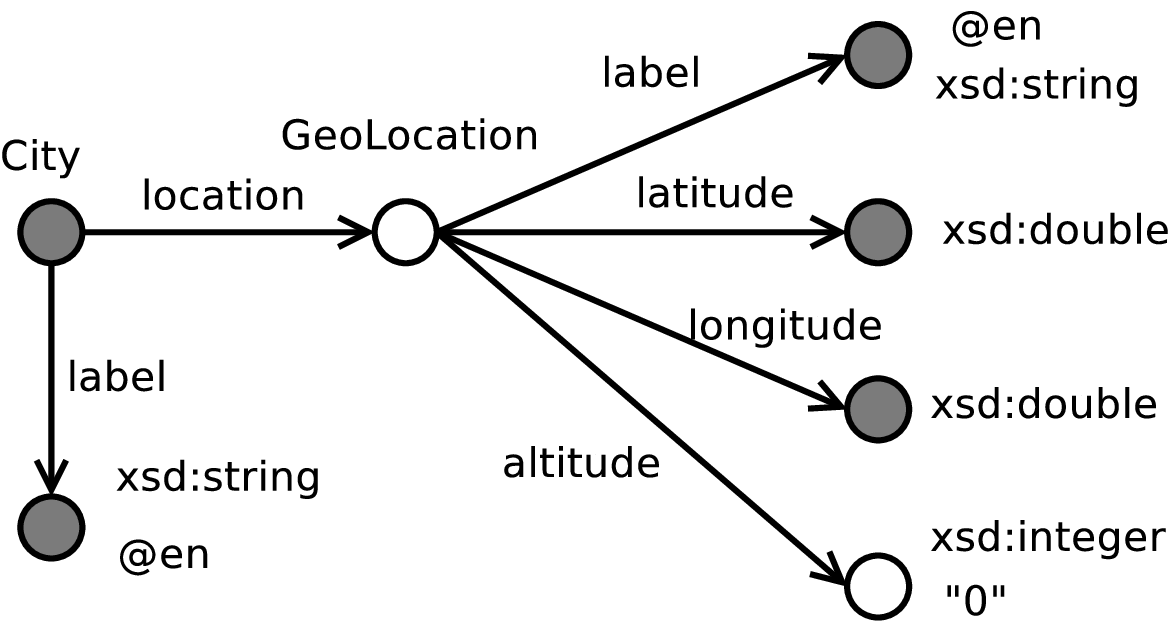}
		\caption{A pattern extracted by \kgmdl{} on the NTNames dataset. Vertices filled in grey are the ports of the pattern.}
		\label{fig:ntnames_cityschema}
	\end{minipage}
\end{figure}

Patterns are not limited to the schema of the data, but they can also present knowledge about the data that is not present in the schema.
While the schema of a dataset describes how facts should be encoded in general, the concrete content of a specific dataset may have some additional structure that stems from the regularity of the real-life facts in the data.

Fig.~\ref{fig:ntnames_cityschema} shows a pattern extracted from the NTNames dataset.
By looking at this pattern we can understand how the geographical information of cities is represented in the data.
But, more importantly, there is an interesting detail in this pattern: the altitude vertex is not a port and has a value of 0, which suggests that cities tend to have an altitude of 0 (so the information is directly encoded in the pattern).
This may seem surprising, but it is actually the case that in this dataset \emph{all} cities have altitude~0, which is probably due to a lack of information.
It would not have been possible to extract this knowledge from the schema alone. Since \kgmdl{}'s patterns are created from the raw facts, they can highlight this type of hidden information that is related to the actual \emph{usage} of properties and classes.

\section{Conclusion}
\label{sec:conclusion}
In this paper we proposed \kgmdl{}, a MDL-based method to extract a human-sized set of descriptive patterns from knowledge graphs and labeled multi-graphs.
In order to do so, we adapted a MDL-based graph mining approach to knowledge graphs.
We showed that the adaptation is not trivial, as KGs have peculiar characteristics that distinguish them from the graphs usually found in the graph mining domain.
We evaluated our approach experimentally on five medium-sized datasets. We showed that it is efficient in finding small sets of patterns that minimize the MDL description length in reasonably short time.
We analyzed the extracted patterns, showing that they allow to infer knowledge both about the schema used to create the data, and the concrete content of the data.
Such patterns may be used to infer the schema from unknown data, create validation constraints (e.g. SHACL shapes), improve the schema by suggesting additions to it, and highlight instances where the raw data is surprising w.r.t. the schema.

As a future work, the approach may be improved to scale better towards large-scale graphs that are common in the semantic web (millions of triples or more).
Because the current approach requires to compute a re-encoded representation of the full data for each candidate sets of patterns in order to compute the MDL description lengths, the scalability is affected.
In order to scale to larger graphs, a sampling approach might be undertaken. However, this is not a trivial matter, as the choice of the sample may affect the extracted patterns.
Therefore the sampled data should be as characteristic as possible w.r.t. the complete data.
Another possible extension is to allow for patterns that contain less-specific matches for literal values, such as substring match or range inclusion.
This would allow for patterns to be potentially insensitive to small changes in literal values, capturing occurrences of slightly different ---but similar--- concrete facts.

\bibliographystyle{plain}
\bibliography{arxiv-kgmdl}

\appendix

\section{SPARQL Queries of Experimental Evaluation Patterns}
\label{sec:appendix:sparql_patterns}

\FloatBarrier
In this section we show the SPARQL queries that correspond to the patterns shown in Section~\ref{sec:experimental_evaluation:qualitative} (we omit the pattern of Fig~\ref{fig:lemon_nounphrases} because of its size).

\begin{figure}[h]
	\centering
	\begin{minipage}{0.4\textwidth}
		\centering
		\includegraphics[width=\textwidth]{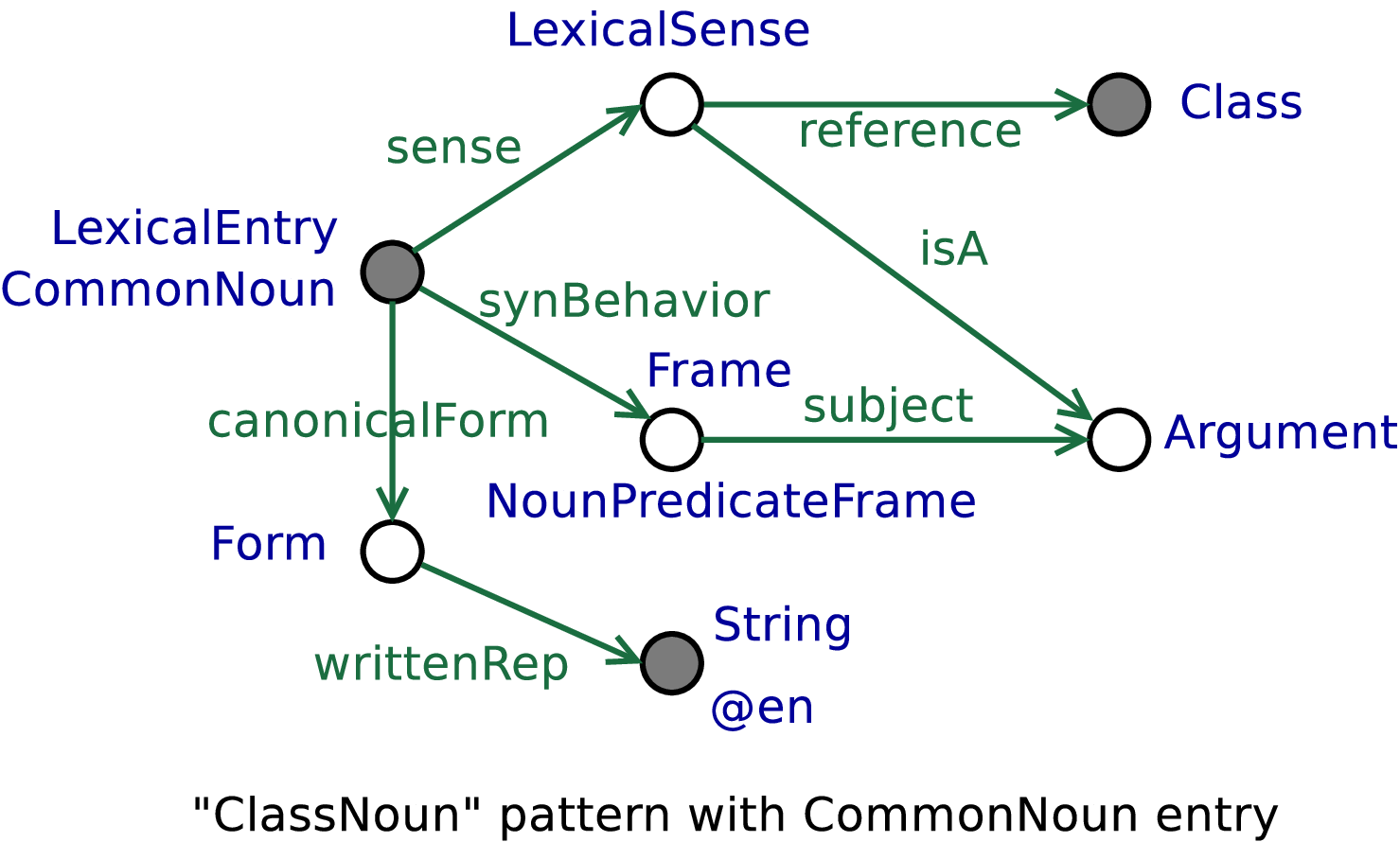}
	\end{minipage}
	\hfil
	\vrule
	\hfil
	\begin{minipage}{0.55	\textwidth}
		\resizebox{\textwidth}{!}{
\begin{lstlisting}
SELECT ?theLexEntry ?theWrittenRep ?theMeaning WHERE {
	?theLexEntry a lemon:LexicalEntry, lexinfo:commonNoun ;
		lemon:canonicalForm [ 
			a lemon:Form ; 
			lemon:writtenRep ?theWrittenRep
		] ;
		lemon:synBehavior [ 
			a lemon:Frame, lexinfo:NounPredicateFrame ;
			lemon:subject ?arg 
		] ;
		lemon:sense [ 
			a lemon:LexicalSense ;
			lemon:isA ?arg ; 
			lemon:reference ?theMeaning
		] .
	?arg a lemon:Argument.
	?theMeaning a owl:Class
	FILTER( lang(?theWrittenRep) = "en" )
}
\end{lstlisting}
		}
	\end{minipage}
	\caption{The SPARQL query corresponding to the pattern of Fig.~\ref{fig:lemon_classnoun}. The chosen query variables are the pattern's ports (in grey in the image).}
\end{figure}

\begin{figure}[h]
	\centering
	\begin{minipage}{0.4\textwidth}
		\centering
		\includegraphics[width=\textwidth]{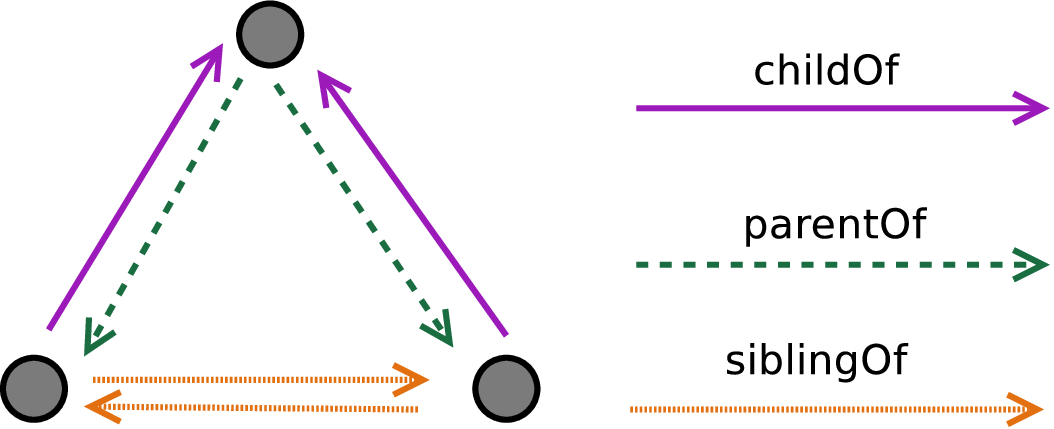}
	\end{minipage}
	\hfil
	\vrule
	\hfil
	\begin{minipage}{0.55	\textwidth}
		\resizebox{\textwidth}{!}{
\begin{lstlisting}
SELECT ?parent ?child1 ?child2 WHERE {
	?parent ntn:parentOf ?child1, ?child2 .
	?child1 ntn:childOf ?parent ; ntn:siblingOf ?child2 .
	?child2 ntn:childOf ?parent ; ntn:siblingOf ?child1 .
}
\end{lstlisting}
		}
	\end{minipage}
	\caption{The SPARQL query corresponding to the pattern of Fig.~\ref{fig:ntnames_siblings}.}
\end{figure}

\begin{figure}[h]
	\centering
	\begin{minipage}{0.4\textwidth}
		\centering
		\includegraphics[width=\textwidth]{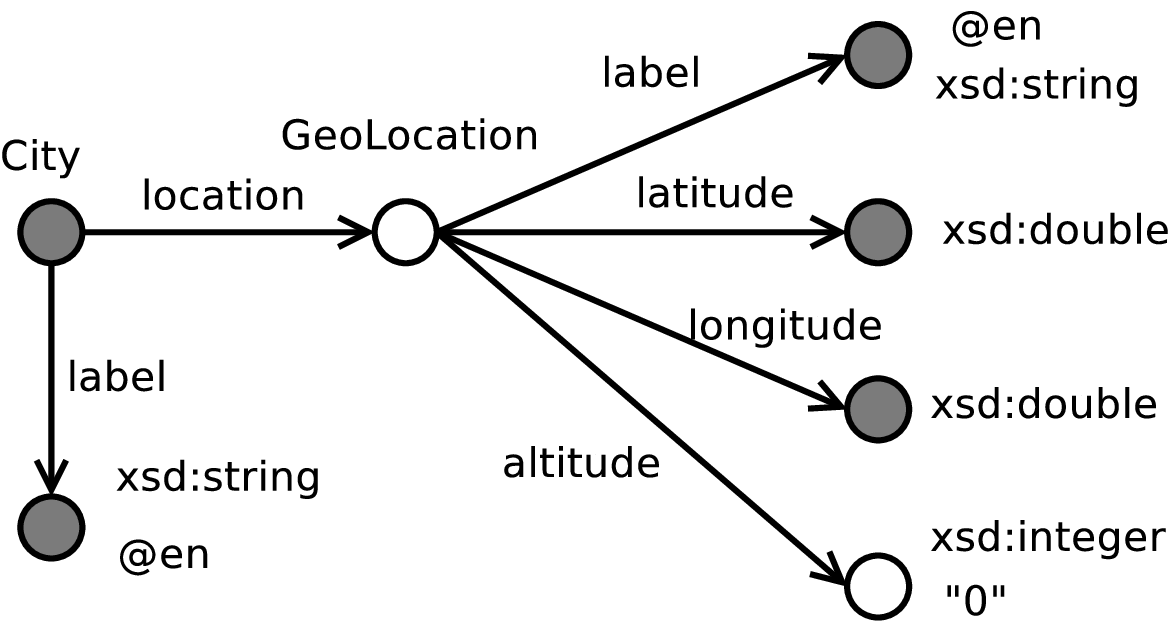}
	\end{minipage}
	\hfil
	\vrule
	\hfil
	\begin{minipage}{0.55	\textwidth}
		\resizebox{\textwidth}{!}{
\begin{lstlisting}
SELECT * WHERE {
	?city rdfs:label ?cityLab ;
		ntn:location ?geoLoc .
	?geoloc	a ntn:GeographicLocation ;
		rdfs:label ?geoLab ;
		ntn:latitude ?lat ;
		ntn:longitude ?long ;
		ntn:altitude "0"^^xsd:integer .
	FILTER	(lang(?cityLab) = "en" && lang(?geoLab) = "en" 
		&& datatype(?lat) = xsd:double 
		&& datatype(?long) = xsd:double)
}
\end{lstlisting}
		}
	\end{minipage}
	\caption{The SPARQL query corresponding to the pattern of Fig.~\ref{fig:ntnames_cityschema}.}
\end{figure}
\FloatBarrier

\section{Prequential Codes}
\label{sec:appendix:prequential}

To compute the MDL description length in Section~\ref{sec:kgmdl:mdl_encoding} we extensively use \emph{prequential plug-in codes}~\cite{mdl_grunwald}.
A prequential plug-in code allows to encode a sequence of items one item at a time, without needing to know the probability of each item in advance (only the possible items need to be known).
Also, prequential codes are asymptotically optimal for large sequences, and ---surprisingly--- do not depend on the order of items in the sequence (which is fundamental when no canonical order exist on a sequence).
In this section we explain how a description length can be computed using prequential codes.

Let ${\cal X}$ be a set of items, and $S = \langle s_0, s_1, \dots, s_{|S|-1} \rangle$ a sequence of items $s_i \in {\cal X}$. We denote $S^i$ the sub-sequence up to and including position $i$ (e.g. $S^2 = \langle s_0, s_1, s_2 \rangle$), and $usage(x, S)$ the number of occurrences of element $x$ in the sequence $S$.
Each element $s_i \in S$ can be encoded optimally using a prefix code based on the distribution of elements before it~\cite{mdl_grunwald}. An initial pseudocount $\epsilon$ is given to each possible element to avoid elements having an empty code (usually $\epsilon = 0.5$ is used in applications).
Therefore the description length for an element $s_i$ based on the previous elements is:
$$L(s_i \mid S^{i-1}) = -\log \left(\frac{usage(s_i, S^{i-1}) + \epsilon}{\sum_{x \in {\cal X}} [usage(x, S^{i-1}) + \epsilon]} \right)$$
Therefore, the description length of the whole sequence is:

\begin{align*}
	L(S) &= \sum_{i=0}^{|S|-1} L(s_i \mid S^{i-1}) \\
		& = \sum_{i=0}^{|S|-1} \left[ -\log \left(\frac{usage(s_i, S^{i-1}) + \epsilon}{\sum_{x \in {\cal X}} [usage(x, S^{i-1}) + \epsilon]} \right) \right ]\\
		\text{\footnotesize \it (sum of logs is log of product)} &= -\log \left( \prod_{i = 0}^{|S|-1} \frac{usage(s_i, S^{i-1}) + \epsilon}{\sum_{x \in {\cal X}} [usage(x, S^{i-1}) + \epsilon]}  \right)
\end{align*}

Since in a product of fractions the order of each operand in the numerator and denominator does not matter, when computing the description length \emph{of a whole sequence} using a prequential code, the order of elements in the sequence does not matter.

Note that in the above expression, the numerator for each occurrence of an element $x \in {\cal X}$ only depends on the previous occurrences \emph{of that same element}, because $usage(x, S)$ yields $\epsilon$ for the first occurrence of $x$, $(1+\epsilon)$ for the second, and so on, independently of the occurrences of other elements. Therefore the numerator can be made clearer by regrouping occurrences of the same element.
The denominator of the above expression only depends on the sum of usages of previous elements (which is the length of $S^{i-1}$ for element $i$), plus a value of $\epsilon$ for each item $x \in {\cal X}$.
Therefore the above expression can be rewritten as:

$$L(S) = -\log\left(
	\frac{
		\prod_{x \in {\cal X}} \prod_{i = 0}^{usage(x, S)-1} (i + \epsilon)
	}{
		\prod_{i=0}^{|S|-1} (i + |{\cal X}|\epsilon)
	}
\right)$$

In the literature, this description length is generally given by employing the gamma function $\Gamma$, which is an extension of factorials to real number: $\Gamma(a) = (a-1)\Gamma(a-1)$ (for an integer $n$, $\Gamma(n) = (n-1)!$).
We can introduce the gamma function in the formula above, by writing out all terms of the products. and using the property that $(n)(n-1)(n-2)\dots(k+1) = \frac{\Gamma(n+1)}{\Gamma(k+1)}$.
We do not detail the whole transformation here, which gives:

\begin{align*}
	L(S) &= -\log \left(
		\frac{\prod_{x \in {\cal X}} \frac{\Gamma(usage(x, S)+\epsilon)}{\Gamma(\epsilon)}}{\frac{\Gamma(|S|+|{\cal X}|\epsilon)}{\Gamma(|{\cal X}|\epsilon)}} \right )\\
		\text{\footnotesize \it (log properties)} &=
			-\sum_{x \in {\cal X}} \left [ \log \left( \frac{\Gamma(usage(x, S) +\epsilon)}{\Gamma(\epsilon)} \right ) \right ]
			+ \log \left( \frac{\Gamma(|S|+|{\cal X}|\epsilon)}{\Gamma(|{\cal X}|\epsilon)} \right )
\end{align*}

Which corresponds to the formulas usually given in the literature.

The advantage of this encoding is that the final formula only depends on the usage of each items~$x \in {\cal X}$, the length of the sequence, and the number of elements of ${\cal X}$. The actual content of the sequence is not needed, which makes the encoding particularly suitable for computing description lengths where the order of elements in the sequence is arbitrary chosen, as the choice does not impact the description length.

\end{document}